\documentclass[10pt,twocolumn,letterpaper]{article}

\usepackage{iccv}
\usepackage{times}
\usepackage{epsfig}
\usepackage{graphicx}
\usepackage{amsmath}
\usepackage{amssymb}
\usepackage{color}
\usepackage{multirow}
\usepackage{amssymb}
\usepackage{algorithm}
\usepackage{algorithmic}
\usepackage{array}

\usepackage[pagebackref=true,breaklinks=true,letterpaper=true,colorlinks,bookmarks=false]{hyperref}

 \iccvfinalcopy 

\def\iccvPaperID{3115} 

\ificcvfinal\fi

\begin{document}

\title{SE\textsuperscript{2}Net: Siamese Edge-Enhancement Network for Salient Object Detection}

\author{Sanping Zhou\textsuperscript{1,3}, Jimuyang Zhang\textsuperscript{3}, Jinjun Wang\textsuperscript{1}, Fei Wang\textsuperscript{2,3}, Dong Huang\textsuperscript{3}\\
	1. Institute of Artificial Intelligence and Robotic, Xi'an Jiaotong University\\
	2. Department of Computer Science, Xi'an Jiaotong University\\
	3. Robotics Institute, Carnegie Mellon University\\
}


\maketitle

\begin{abstract}
Deep convolutional neural network significantly boosted the capability of salient object detection in handling large variations of scenes and object appearances. However, convolution operations seek to generate strong responses on individual pixels, while lack the ability to maintain the spatial structure of objects. Moreover, the down-sampling operations, such as pooling and striding, lose spatial details of the salient objects. In this paper, we propose a simple yet effective Siamese Edge-Enhancement Network~(SE\textsuperscript{2}Net) to preserve the edge structure for salient object detection. Specifically, a novel multi-stage siamese network is built to aggregate the low-level and high-level features, and parallelly estimate the salient maps of edges and regions. As a result, the predicted regions become more accurate by enhancing the responses at edges, and the predicted edges become more semantic by suppressing the false positives in background. After the refined salient maps of edges and regions are produced by the SE\textsuperscript{2}Net, an edge-guided inference algorithm is designed to further improve the resulting salient masks along the predicted edges. Extensive experiments on several benchmark datasets have been conducted, which show that our method is superior than the state-of-the-art approaches~\footnote{In the near future, we will release our codes for the public research.}.
\end{abstract}

\section{Introduction}
\label{sec:introd}
Salient object detection aims at identifying the visually interesting object regions that are consistent with human perception. This algorithm has been a fundamental module in many visual tasks, such as image retrieval~\cite{Babenko_Lempitsky:2015}, object tracking~\cite{Pu_Song_Ma:2018}, scene classification~\cite{Zhang_Du_Zhang:2015}, video segmentation~\cite{Wang_Shen_Porikli:2015}, action detection~\cite{Yeung_Russakovsky_Mori:2016} and etc. With the powerful non-linearity learning nature of Deep Neural Network~(DNN), significant progresses have been made in salient object detection. Most methods try to directly learn salient mappings from a raw input image to a heatmap of salient region in an end-to-end manner~\cite{Wang_Lu_Wang:2017,Wang_Borji_Zhang:2017,Zhu_Liang_Wei:2014}. The key advances made by DNN are in the capability of salient object detection in handling large variations of scenes and object appearances.  

However, deep convolutional operations in DNN seek to generate strong responses on individual pixels, while lack the ability to maintain the spatial structure of objects.  The mainstream methods~\cite{Deng_Hu_Zhu:2018,Liu_Han_Yang:2018,Liu_Qiu_Zhang:2018,Ronneberger_Fischer_Brox:2015,Wang_Zhang_Wang:2018} have extensively studied how to fuse the low-level and high-level features in the past few years, so as to jointly improve the precision of regional localization and enhance the pixel-wise response of salient maps. U-Net~\cite{Ronneberger_Fischer_Brox:2015} is a typical network structure that aggregates the low-level and high-level features. Moreover, the estimated salient maps are still very blurred in detailed structures, especially at edges, due to the down-sampling operations, such as pooling and striding. As a fine-grained spatial structure in salient maps, edges are critical to be estimated in salient object detection. 

\begin{figure}[t]
	\centering
	\begin{tabular}{c}
		\hspace{-0.2cm}
		\includegraphics[height = 4.0cm, width = 8.1cm]{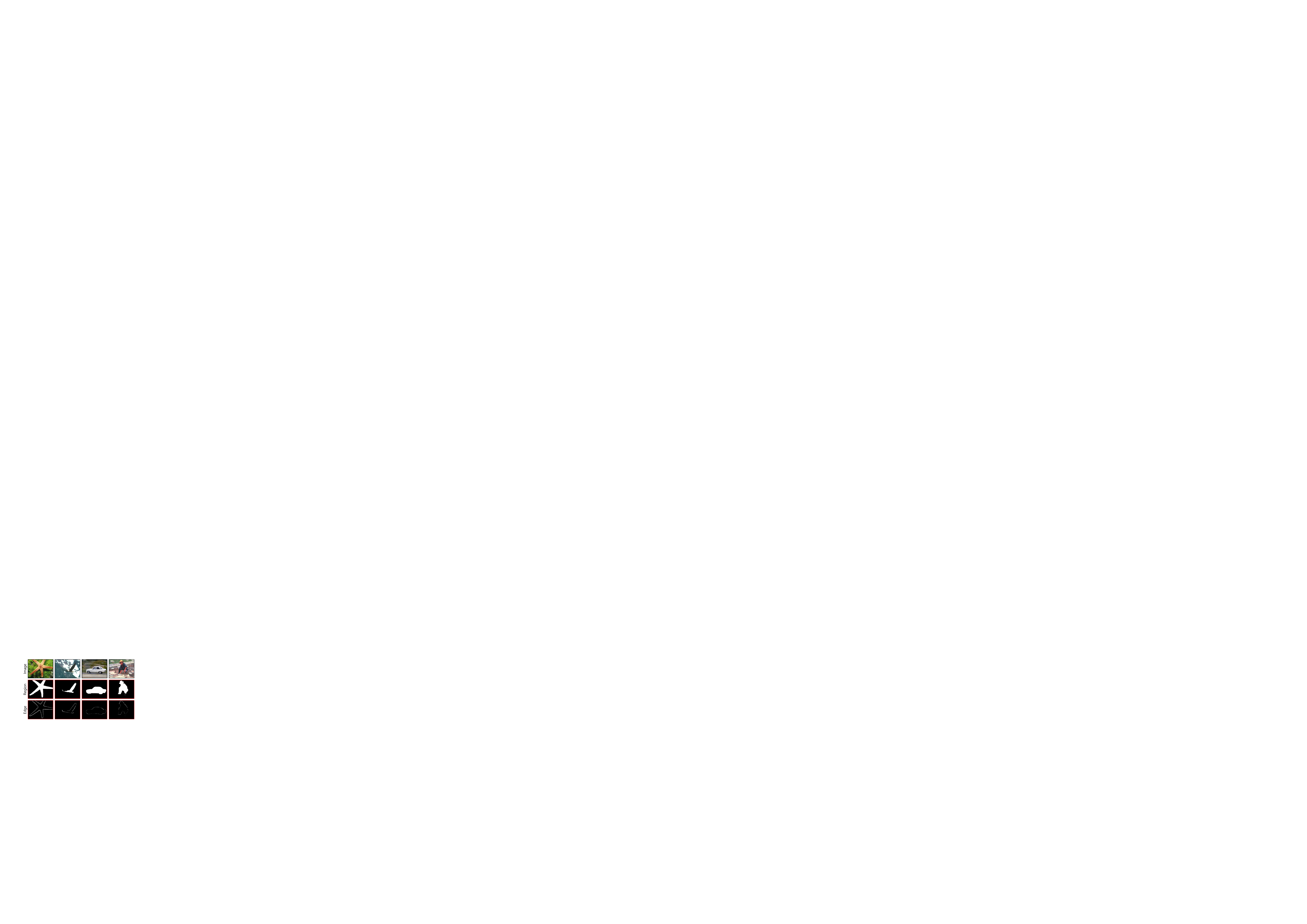}
	\end{tabular}
	\caption{Illustration of some salient object detection examples on the ESSCD dataset~\cite{Yan_Xu_Shi:2013} by our SE\textsuperscript{2}Net, in which the first to third columns represent the input images, the estimated region maps, and the estimated edge maps, respectively.}
	\label{fig_1}
	\vspace{-0.2cm}
\end{figure}

To our best knowledge, pioneering works~\cite{Hou_Cheng_Hu:2017,Li_Yu:2015,Li_Yu:2016,Wang_Lin_Lu:2015,Zhu_Liang_Wei:2014,Wang_Zhang_Wang:2018,Wang_Wang_Lu:2016,Luo_Mishra_Achkar:2017,Zhang_Wang_Lu:2017,Li_Xie_Lin:2017} have already been exploring how to enhance edges in learning the salient maps of regions. Their main solution is to use the discontinuous property of edges to improve the predicted masks along edges. By incorporating the prior knowledge on edges, those methods can preserve the object edges in a weakly-supervised manner. However, the precision of edges may be seriously degenerated, because it is hard to compute the precise location of edges from the semantic region masks. In the salient object detection task, we argue that it is more suitable to detect the object edges in a fully-supervised manner, because: (1) the discontinuities of object edges can't provide enough information to solve the dense tasks, such as the edge detection~\cite{Yang_Proce_Cohen:2016} and instance segmentation~\cite{Ren_Zemel:2017}; and (2) the ground truth of edges can be easily obtained from the ground truth of masks by using the off-the-shelf edge detectors, such as Canny~\cite{Gao_Zhang_Yang:2010} and Laplacian~\cite{Van_Pieter:2001}. Based on these understandings, we aim to learn the salient maps of edges and regions in an end-to-end network.

In this paper, we propose a simple yet effective Siamese Edge-Enhancement Network~(SE\textsuperscript{2}Net) that can jointly estimate the salient maps of edges and regions, as shown in Figure~\ref{fig_1}. To achieve high-quality salient maps, we build SE\textsuperscript{2}Net as a novel multi-stage siamese network that aggregates the low-level and high-level features. Each stage of the network takes the estimations of edges and regions from the previous stage as inputs. Then new estimations of current stage, along with the low-level and high-level features, are fed into the next stage to refine the previous salient estimations. With the number of stages increases, the estimated salient maps of edges and regions will be gradually refined until a remarkable improvement is achieved. Last but not least, a novel edge-guided inference algorithm is designed to further improve the resulting masks along the predicted edges. The detailed defects near the predicted edges are effectively corrected after this refinement procedure. 

The main contributions of this work can be highlighted as follows: 1) A novel multi-stage siamese network is designed to jointly estimate the edges and regions from the low-level and high-level features, in which the region sub-network and edge sub-network can learn from each other in a complementary way. 2) A novel edge-guided inference algorithm is designed to improve the resulting masks along the predicted edges, which is very effective and efficient in refining the final predictions. Extensive experimental results on several public datasets, including the DUTS~\cite{Wang_Lu_Wang:2017}, ECSSD~\cite{Yan_Xu_Shi:2013}, SOD~\cite{Movahedi_Elder:2010}, DUT-OMRON~\cite{Yang_Zhang_Lu:2013}, THUR 15K~\cite{Cheng_Mitra_Huang:2014} and HKU-IS~\cite{Li_Yu:2015}, show that our method has achieved a significant improvement as compared with the state-of-the-art approaches.

\begin{figure*}[t]
	\centering
	\begin{tabular}{c}
		\includegraphics[height = 6.0cm, width = 17.0cm]{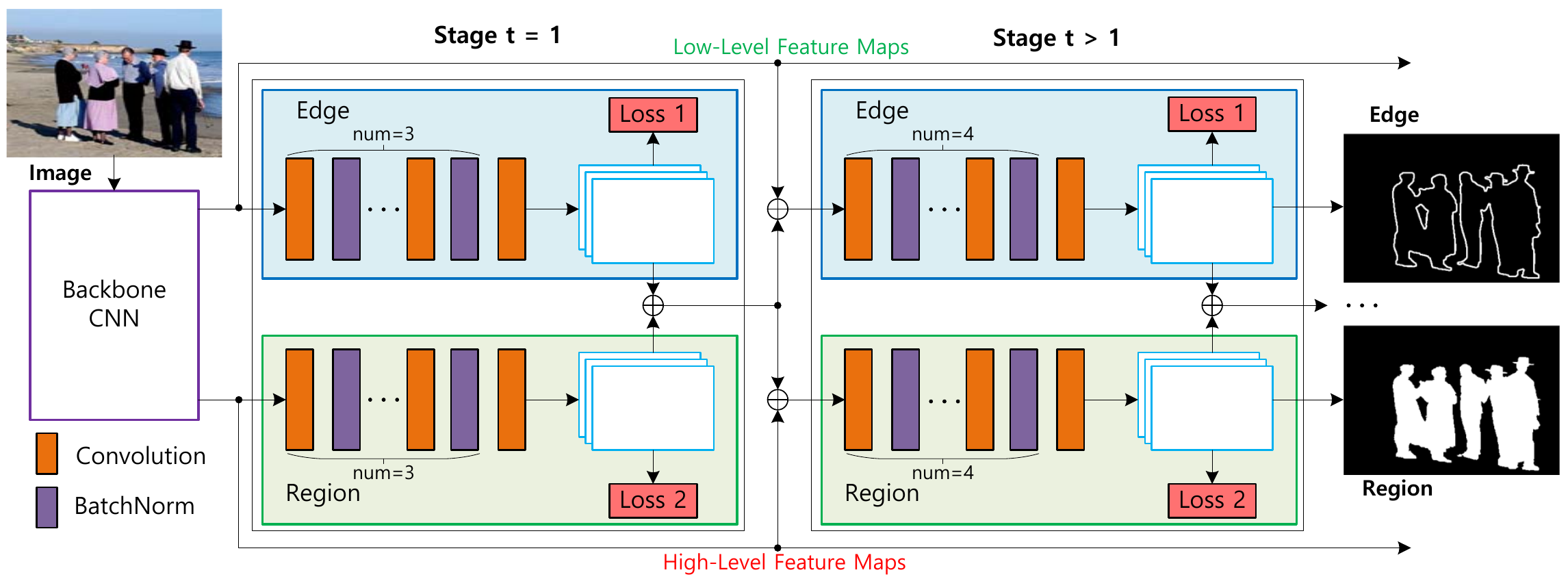}
	\end{tabular}
	\caption{Illustration of our SE\textsuperscript{2}Net. Firstly, we generate the low-level and high-level features through a backbone network. Secondly, the initial salient maps of edges and regions are learned from the low-level and high-level features in the first stage, respectively. Thirdly, the predicted maps of edges and regions along with the low-level and high-level features are used to refine the previous predictions in the subsequent stages.}
	\label{fig_2}
	\vspace{-0.2cm}
\end{figure*}

\section{Related Work}
We review two lines of related works, {\em i.e.}, the salient object detection and salient edge detection, in the following paragraphs.

\textbf{Salient object detection.} The early salient object detection methods usually integrate different kinds of features and prior knowledge to model the focused attention of human beings. For example, Cheng et al.~\cite{Cheng_Mitra_Huang:2015} introduced a region contrast based algorithm to consider the spatial coherence across the neighboring regions and the global contrast over the entire image, simultaneously. In~\cite{Li_Lu_Zhang:2013}, Li et al. proposed the dense and sparse reconstruction errors based on the background prior for salient object detection. Different from these traditional methods, later researchers focus on how to learn a consistent salient mapping function from the raw input images~\cite{Wang_Borji_Zhang:2017,Wang_Lu_Wang:2017,Liu_Han_Yang:2018}, which can achieve very promising results with the help of DNN. In early works, this line of methods only utilize deep features to predict the saliency scores of image segments, such as superpixels or object proposals. For instance, Wang et al.~\cite{Wang_Lu_Ruan:2015} proposed two convolutional neural networks~(CNN) to combine the local superpixel estimation and global proposal search for salient object detection. In~\cite{Li_Yu:2015}, Li et al. computed the saliency value of each superpixel by extracting its contextual CNN features. In recent years, more works have been focused on designing an end-to-end framework to learn the salient maps from raw input images. For example, the skip connection~\cite{Ronneberger_Fischer_Brox:2015,Li_Yu:2016,Hou_Cheng_Hu:2017,Zhang_Wang_Lu:2017} has been extensively studied to fuse the low-level and high-level features, so as to improve the accuracy of predicted salient regions. In~\cite{Deng_Hu_Zhu:2018,Kuen_Wang_Wang:2016,Liu_Han:2016,Wang_Wang_Lu:2016}, different recurrent structures have been carefully designed to reduce the prediction errors by iteratively integrating the contextual information.

\textbf{Salient edge detection.} The traditional edge detectors, such as Canny~\cite{Gao_Zhang_Yang:2010} and Laplacian~\cite{Van_Pieter:2001}, are designed based on the gradient information. Therefore the edge predictions are very noisy when the input images contain many complex structures. In the salient edge detection, the models need to jointly emphasize the visually interesting edges while suppress the uninteresting ones~\cite{Liu_Cheng_Hu:2017}, therefore it is more challenging than the traditional edge detection. To address this challenge, many deep learning based methods have been developed to extract rich features for salient edge detection. The first line of methods try to design different network structures to make full use of the low-level and high-level features~\cite{Bertasius_Shi_Torresani:2015}. For example, Xie and Tu~\cite{Xie_Tu:2015} designed a holistically-nested edge detection network to learn hierarchical features to tell the ambiguity between natural image edge and object boundary detection. In~\cite{Yang_Price_Cohen:2016}, Yang et al. introduced an encoder-decoder network to predict the higher-level object contours. The second line of methods aim to impose different learning strategies and loss functions to guide the training process~\cite{Kokkinos:2015}. For instance, Shen et al.~\cite{Shen_Wang_Wang:2015} proposed a novel positive-sharing loss function to enable that each subclass can share the same loss for the whole positive class. In~\cite{Liu_Lew:2016}, Liu and Lew introduced a relaxed deep supervision signal by merging the detection results of Canny and the general ground truth to conduct the coarse-to-fine edge detection. In recent years, the object classes together with salient annotations~\cite{Yu_Feng_Liu:2017,Liu_Cheng_Bian:2018} have been extensively used to promote the salient edge detection into semantic edge detection.

Considering the fact that the salient object detection and salient edge detection can benefit from each other in the training process, we incorporate them in a unified network which can enhance the responses at edges for salient object prediction and suppress the false positives in background for salient edge prediction, simultaneously.

\section{Siamese Edge Enhancement Network}
Denote the training dataset  as $\mathbf{Y} = \{\mathbf{X}_i, \mathbf{R}_i^*, \mathbf{E}_i^*\}_{i=1}^N$, where $\mathbf{X}_i$ indicates the $i^{th}$ raw input image, $\mathbf{R}_i^* \in  \mathbb{R}^{w \times h}$ and $\mathbf{E}_i^* \in  \mathbb{R}^{w \times h}$ represent the corresponding labels of salient regions and edges, and $N$ is the number of training samples.  Our ultimate goal is to enhance weak edges in the predicted masks. To achieve this goal, we propose two novel approaches, {\em i.e.}, a multi-stage siamese network and an edge-guided inference algorithm, which are explained in details in the following paragraphs.
\subsection{Multi-stage Siamese Network}
As shown in Figure.~\ref{fig_2}, our multi-stage siamese network can jointly estimate the salient maps of edges and regions. The network consists of two branches, {\em i.e.}, the edge branch and region branch, which are in the same structure but do not share any parameter. The edge branch predicts the salient maps of edges from the low-level features, and the region branch estimates the salient maps of regions from the high-level features. Each branch is an iterative prediction architecture, which can refine the predictions over successive stages, $t \in \{1, \cdots, T\}$, under the supervision of ground truth at each stage. Specifically, we directly estimate the salient maps of edges and regions from the low-level and high-level features at the first stage. Once the salient maps of edges and regions are predicted, we concatenate them together to generate a new salient region map which has a strong response at the edges. Then, the new generated salient map along with the low-level and high-level features are further fed into the next stage to predict the better salient maps of edges and regions, respectively. The detailed structure at each stage is summarized in Table~\ref{tab_1}, which is mainly consisted of three different layers: the normal convolution layer, batch normalization layer~\cite{Ioffe_Szegedy:2015} and parametric rectified linear unit~\cite{He_Zhang_Ren:2015}.

In the training process, we first pass each image through a backbone network, {\em i.e.}, VGG16~\cite{Simonyan_Zisserman:2014}, ResNet50~\cite{He_Zhang_Ren:2016} or ResNext101~\cite{Xie_Girshick_Tu:2017}, to generate a set of feature maps. As a result, five scales of feature maps, namely $1$, $1/2$, $1/4$, $1/8$ and $1/16$ of the input size, are computed to generate the low-level and high-level features. In particular, the first three scales of feature maps are concatenated to generate the low-level features $\mathbf{L}_i$, and the last two scales of feature maps are concatenated to generate the high-level features $\mathbf{H}_i$. Then, the salient maps of edges and regions at the first stage can be represented by $\mathbf{E}_i^1 = \psi_e^1(\mathbf{L}_i)$ and  $\mathbf{R}_i^1 = \psi_r^1(\mathbf{H}_i)$, where $\psi_e^1(\cdot)$ and $\psi_r^1(\cdot)$ denote the edge mapping function and region mapping function at the first stage, respectively. In each subsequent stage, the predictions of edges and regions in the previous stage, along with the low-level and high-level features, will be concatenated together and further used to refine the previous predictions. This process can be formulated as follows:
\begin{equation}
\label{eq_1}
\mathbf{E}_i^t = \psi_e^t(\mathbf{L}_i, \mathbf{E}_i^{t-1}, \mathbf{R}_i^{t-1}), \forall t > 1,
\end{equation} 
\vspace{-0.3cm}
\begin{equation}
\label{eq_2}
\mathbf{R}_i^t = \psi_r^t(\mathbf{H}_i, \mathbf{E}_i^{t-1}, \mathbf{R}_i^{t-1}), \forall t > 1,
\end{equation}
where $\psi_e^t(\cdot)$ and $\psi_r^t(\cdot)$ denote the edge mapping function and region mapping function at stage $t$, respectively. Besides, the $\mathbf{E}_i^{t-1}$ and $\mathbf{R}_i^{t-1}$ represent the resulting salient maps of edges and regions at stage $t-1$.

\begin{table}[t]
	\centering
	\small
	\begin{tabular}{ | p{1cm}<{\centering} | p{1.5cm}<{\centering} |  p{2.5cm}<{\centering} |  p{1.5cm}<{\centering} |}
		\hline
		\textbf{Stages} & \textbf{Layers} & \textbf{Kernels} &  \textbf{Padding}\\
		\hline
		\multicolumn{1}{|c|}{\multirow{9}{*}{$t = 1$}}
		& Conv & $3 \times 3\times 256$ & 1\\
		\cline{2-4}
		& \multicolumn{3}{c|}{BN + PReLU} \\
		\cline{2-4}
		& Conv & $3 \times 3\times 256$ & 1\\
		\cline{2-4}
		& \multicolumn{3}{c|}{BN + PReLU} \\
		\cline{2-4}
		& Conv & $3 \times 3\times 256$ & 1\\
		\cline{2-4}
		& \multicolumn{3}{c|}{BN + PReLU} \\
		\cline{2-4}
		& Conv & $1 \times 1\times 512$ & 0\\
		\cline{2-4}
		& \multicolumn{3}{c|}{BN + PReLU} \\
		\cline{2-4}
		& Conv & $1 \times 1\times 1$ & 0\\
		\hline
		\multicolumn{1}{|c|}{\multirow{11}{*}{$t > 1$}}
		& Conv & $7 \times 7\times 128$ & 3\\
		\cline{2-4}
		& \multicolumn{3}{c|}{BN + PReLU} \\
		\cline{2-4}
		& Conv & $7 \times 7\times 128$ & 3\\
		\cline{2-4}
		& \multicolumn{3}{c|}{BN + PReLU} \\
		\cline{2-4}
		& Conv & $7 \times 7\times 128$ & 3\\
		\cline{2-4}
		& \multicolumn{3}{c|}{BN + PReLU} \\
		\cline{2-4}
		& Conv & $7 \times 7\times 128$ & 3\\
		\cline{2-4}
		& \multicolumn{3}{c|}{BN + PReLU} \\
		\cline{2-4}
		& Conv & $1 \times 1\times 128$ & 0\\
		\cline{2-4}
		& \multicolumn{3}{c|}{BN + PReLU} \\
		\cline{2-4}
		& Conv & $1 \times 1\times 1$ & 0\\
		\hline
	\end{tabular}
	\caption{The detailed network structure at each stage, in which `Conv' means the normal convolution layer, `BN' denotes the batch normalization layer and `PReLU' indicates the parametric rectified linear unit.}
	\label{tab_1}
\end{table}

\begin{figure}[t]
	\centering
	\begin{tabular}{c}
		\hspace{-0.3cm}
		\includegraphics[height = 2.4cm, width = 8.3cm]{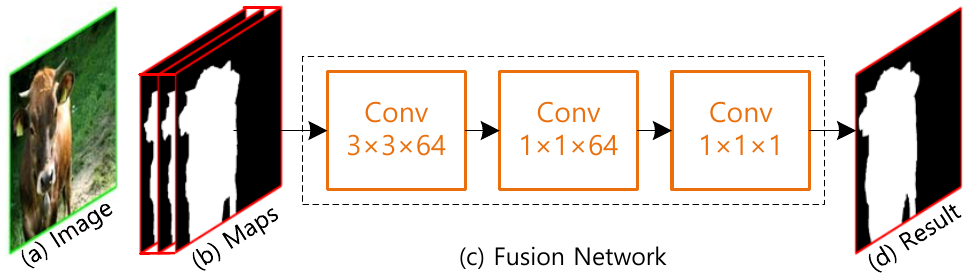}
	\end{tabular}
	\caption{Illustration of our fusion network. Specifically, we first concatenate the salient maps at each stage together, afterwards the obtained tensors are further passed through a three-layer fusion network to obtain the final prediction results.}
	\label{fig_3}
	\vspace{-0.2cm}
\end{figure}

Given a new testing image, our siamese multi-stage network can predict a set of salient maps of edges $\mathbf{E}_i = \{\mathbf{E}_i^1, \cdots, \mathbf{E}_i^T\}$ and regions $\mathbf{R}_i = \{\mathbf{R}_i^1, \cdots, \mathbf{R}_i^T\}$ at the outputs of all stages. In general, the quality of salient maps is consistently improved over the stages, therefore one can directly take predictions at the last stage as the final results. What's more, we further design a simple yet effective fusion network to fuse the predictions from all stages.  The final salient maps of edges and regions can be simply formulated as follows:
\begin{equation}
\label{eq_3}
\mathbf{E}_i^0 = \varphi_e^0(\mathbf{E}_i^1, \cdots, \mathbf{E}_i^T),
\end{equation} 
\vspace{-0.3cm}
\begin{equation}
\label{eq_4}
\mathbf{R}_i^0 = \varphi_r^0(\mathbf{R}_i^1, \cdots, \mathbf{R}_i^T),
\end{equation}
where $\varphi_e^0(\cdot)$ and $\varphi_r^0(\cdot)$ denote the edge fusion function and region fusion function, respectively. As shown in Figure~\ref{fig_3}, our fusion network is a three-layer CNN: The first convolutional layer has a kernel in size of $3\times3\times64$, which is used to learn more local information in a small neighborhood. The size of second convolutional layer is $1\times1\times64$, which is designed to weight pixels across channels. The third convolutional layer is in size of $1\times1\times1$, which is used to generate the final salient map of edges or regions. In practice, the fusion network and our multi-stage siamese network are trained in an end-to-end manner. 

Up to now, we can learn the salient maps of edges and regions under the supervision of ground truth at each stage. To train our SE\textsuperscript{2}Net, we introduce two weighted $\mathrm{L}2$ loss functions to minimize the errors between the predictions and ground truths, which are defined as follows:
\begin{equation}
\label{eq_5}
\mathcal{E}_i^t = \sum\limits_{x \in \mathbf{E}_i^t} \sum\limits_{y \in \mathbf{N}_x} K_\sigma (x-y) \|\mathbf{E}_i^t(x) - \mathbf{E}_i^*(y)\|_F^2, \forall t \geq 0,
\end{equation} 
\vspace{-0.2cm}
\begin{equation}
\label{eq_6}
\hspace{-0.05cm}
\mathcal{R}_i^t = \hspace{-0.1cm} \sum\limits_{x \in \mathbf{R}_i^t} \sum\limits_{y \in \mathbf{N}_x} K_\sigma(x-y) \|\mathbf{R}_i^t(x) - \mathbf{R}_i^*(y)\|_F^2, \forall t \geq 0,
\end{equation} 
where $K_\sigma(x-y)$ represents a truncated Gaussian kernel with the standard deviation of $\sigma$, which can be formulated as follows:
\begin{equation}
\label{eq_7}
K_\sigma(x-y) = \hspace{-0.1cm} \left\{
\begin{split}
\hspace{-0.1cm} \frac{1}{\sqrt 2\pi \sigma}\exp (- \frac{|x - y|^2}{2 \sigma ^2} )&, if \hspace{0.05cm}|x \hspace{-0.05cm}-\hspace{-0.05cm} y| \le \rho, 
\\\hspace{-0.1cm} 0 \hspace{1.2cm}&,\hspace{0.05cm}else.
\end{split} \right.\hspace{-0.1cm},\hspace{-0.1cm}
\end{equation}
where $\rho$ indicates the radius of local neighborhood $\mathbf{N}_x$ which is centered at the point of $x$. The advantage of our weighted $\mathrm{L}2$ loss function over the standard one is that, it considers the regression problem in a local neighborhood, therefore the learned maps are robust to the salient object annotations.

Finally, we extend our weighted $\mathrm{L}2$ loss function into all the training samples and all the network stages, then the overall objective function can be formulated as follows:
\begin{equation}
\label{eq_8}
\mathcal{J} = \frac{1}{N \times (T + 1)} \sum\limits_{i = 1}^ N \sum\limits_{t = 0}^ T \mathcal{E}_i^t + \mathcal{R}_i^t. 
\end{equation} 

\subsection{Edge-guided Inference Algorithm}
Although the DNN based methods can usually obtain the high-quality masks of salient objects, the resulting salient maps may be not very smooth or precise at the output layers. Therefore, post-processing algorithms are usually needed to further boost the final results.

As done in most of the salient object detection methods, we first apply the fully connected conditional random field~(CRF) algorithm~\cite{Krahenbuhl_Koltun:2011} to refine the final results during the inference phase. The energy function of CRF is defined as follows:
\begin{equation}
\label{eq_9}
E(s) =  \sum\limits_{i} \theta_i(s_i) + \sum\limits_{i,j} \theta_{ij}(s_i, s_j),
\end{equation} 
where $s$ denotes the initial prediction results. The unary potential $\theta_i$ is computed independently for each pixel using the initial response value of pixel $s_i$, and the pairwise potential $\theta_{ij}$ encourages similar responses for pixels in similar values and close spatial details.

\begin{figure}[t]
	\centering
	\begin{tabular}{c}
		\hspace{-0.2cm}
		\includegraphics[height = 2.8cm, width = 8.0cm]{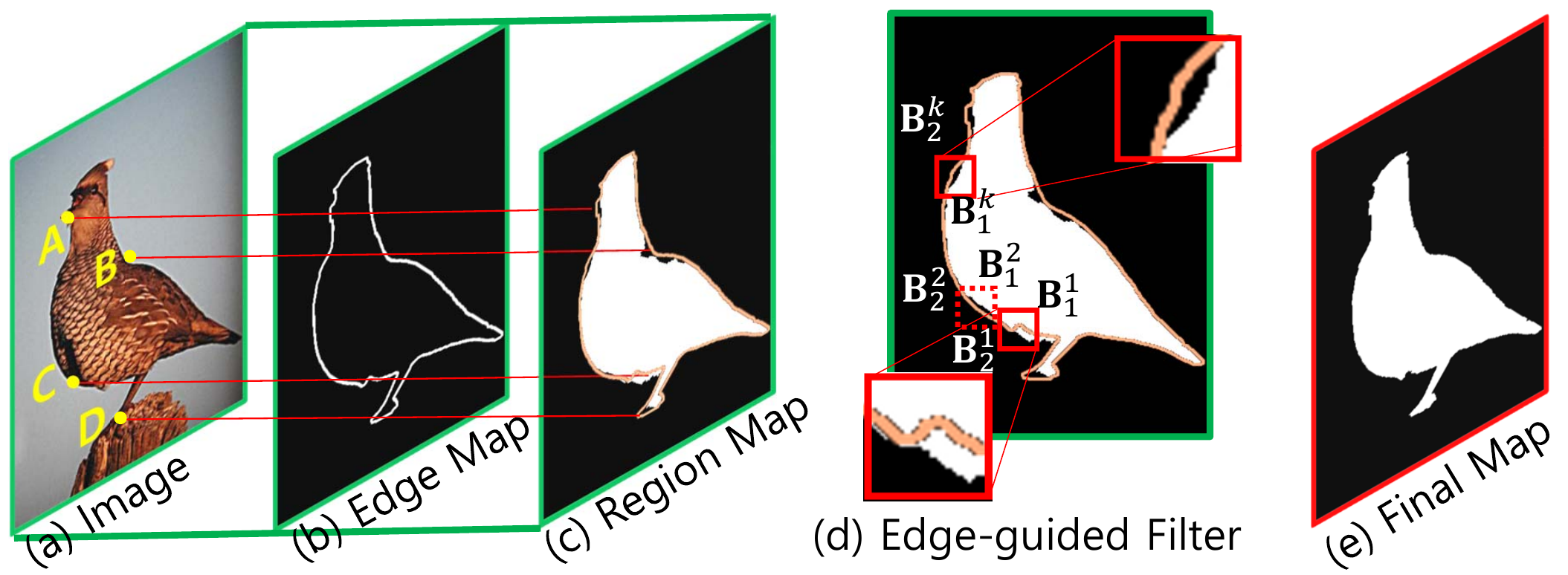}
	\end{tabular}
	\caption{Illustration of our edge-guided inference algorithm, in which the salient maps of edges and regions are firstly refined by the CRF based method in (b) and (c), then the edge-guided filter is used to refine the regions along the predicted edges.}
	\label{fig_4}
	\vspace{-0.2cm}
\end{figure}

Novel to most of the salient object detection methods, since our network can jointly predict the salient maps of edges and regions, we developed a novel edge-guided inference algorithm to filter small bumpy regions along the predicted edges. The main idea is shown in Figure~\ref{fig_4}, in which the final salient maps of regions may still contain some small defects along the predicted edges, denoted by points A, B, C and D in Figure~\ref{fig_4}~(a). As shown in Figure~\ref{fig_4}~(e), these defects can be effectively corrected by our edge-guided inference algorithm. In particular, we first generate a set of successive rectangle boxes $\mathbf{B} = \{\mathbf{B}^k\}_{k=1}^K$ to cover the predicted edges, in which each rectangle box $\mathbf{B}^k$ is in size of $5\times5$ and shares the same center with edges, besides $\mathbf{B}^{k-1} \cap \mathbf{B}^{k} = \emptyset$ while $\mathbf{B}^{1} \cap \mathbf{B}^{K} \neq \emptyset$. Then, we can find that the predicted edges divide each rectangle box into two parts $\mathbf{B}^k = \{\mathbf{B}_1^k, \mathbf{B}_2^k\}$, in which the salient and unsalient regions usually occupy different sizes in each part. If the salient region is larger than the non-salient region, there is a high probability that the whole part represents the salient region, and vice versa. Based on this observation, we can filter each pixel $(x,y)$ in $\mathbf{B}_j^k$ as follows:
\begin{equation}
\label{eq_10}
\mathbf{B}_j^k(x,y) = \left\{
\begin{split}
1&, \hspace{0.1cm}if \hspace{0.1cm} \eta_j^k > 1, \\
0&, \hspace{0.1cm}else.
\end{split} \right. ,
\end{equation}
where $\eta_j^k$ = {\footnotesize $\frac{\mathcal{A}(\mathbf{B}_j^k = 1)}{\mathcal{A}(\mathbf{B}_j^k = 0)}$} denotes the ratio between salient region

{
\vspace{-0.15cm}
\noindent
and non-salient region in the $j^{th}$ part of the $k^{th}$ rectangle box, and $\mathcal{A}(\cdot)$ indicates the area operator. We conclude the overall inference process in Algorithm~\ref{alg}. Because we further use the edge information to help refine the region masks, we name it as the edge-guided inference algorithm.}

\section{Experiments}
\subsection{Experimental Settings}
\textbf{Dataset.} We used the DUTS dataset~\cite{Wang_Lu_Wang:2017} to train our SE\textsuperscript{2}Net network. The DUTS dataset is a latest released challenging dataset that contains $10,553$ training images and $5,019$ testing images in very complex scenarios. As indicated in~\cite{Hou_Cheng_Hu:2017}, a good salient object detection model should work well over almost all datasets, therefore we also evaluated our model on the other five datasets, {\em i.e.}, the ECSSD~\cite{Yan_Xu_Shi:2013}, SOD~\cite{Movahedi_Elder:2010}, DUT-OMRON~\cite{Yang_Zhang_Lu:2013}, THUR 15K~\cite{Cheng_Mitra_Huang:2014} and HKU-IS~\cite{Li_Yu:2015}, which contains $1,000$, $300$, $5,168$, $6,232$, and $4,447$ natural images, respectively. In each image, there are different numbers of salient objects with diverse locations. For fair comparison, we follow the same data partition as in~\cite{Liu_Qiu_Zhang:2018}.

Our SE\textsuperscript{2}Net requires the annotations of both edges and regions, while the existing datasets can only provide the ground truth of regions. We generate the ground truth of edges in a very simple and cheap two-step approach: (1) Generate the edge annotations from the ground truth of regions by using the Canny~\cite{Gao_Zhang_Yang:2010} operator; (2) Dilate the width of each edge annotation to five pixels.

\textbf{Evaluation Metric.} We used two metrics~\footnote{Their definitions can be found in~\cite{Liu_Qiu_Zhang:2018}, and we set $\beta^2 = 0.3$ for fair comparison.} to quantitatively evaluate the edges and regions, {\em i.e.}, the F-measure score~($\mathrm{F}_\beta$) and Mean Absolute Error~($\mathrm{MAE}$) in the testing phase. In the following comparison, a better salient object detection model should have a larger $\mathrm{F}_\beta$ and a smaller $\mathrm{MAE}$.

\begin{algorithm}[t]
	\caption{Edge-guided inference algorithm}
	\label{alg}
	\begin{algorithmic}[1]
		\STATE {\bfseries Input:} Inital maps: $\mathbf{E} = \{\mathbf{E}_i\}_{i=1}^N$ and $\mathbf{R} = \{\mathbf{R}_i\}_{i=1}^N$.
		\STATE {\bfseries Output:} Refined maps: $\mathbf{R}^{\dagger} = \{\mathbf{R}_i^{\dagger}\}_{i=1}^N$.
		\FOR{$i=1$; $i<N$; $i+\hspace{-0.1cm}+$}
		\STATE $\mathbf{E}_i = \mathrm{CRF}(\mathbf{E}_i)$;
		\STATE $\mathbf{R}_i = \mathrm{CRF}(\mathbf{R}_i)$;
		\STATE Generate a set of rectangle boxes $\mathbf{B} = \{\mathbf{B}^k\}_{k=1}^K$;
		\FOR{$k = 0$; $k < K$; $k+\hspace{-0.1cm}+$}
		\FOR{$(x,y) \in \mathbf{B}_j^j$}
		\IF{$\eta_j^k > 1$}
		\STATE $\mathbf{B}_j^k(x,y) = 1$;
		\ELSE
		\STATE $\mathbf{B}_j^k(x,y) = 0$;
		\ENDIF
		\ENDFOR		
		\ENDFOR
		\STATE Update the $\mathbf{R}_i^{\dagger}$ using both $\mathbf{B}$ and $\mathbf{R}_i$;
		\ENDFOR
	\end{algorithmic}
\end{algorithm}

\begin{table*}[t]
	\centering
	\small
	\begin{tabular}{| p{1.5cm} | p{0.85cm}<{\centering} | p{0.85cm}<{\centering} | p{0.85cm}<{\centering} | p{0.85cm}<{\centering} | p{0.85cm}<{\centering} | p{0.85cm}<{\centering} | p{0.85cm}<{\centering} | p{0.85cm}<{\centering} | p{0.85cm}<{\centering} | p{0.85cm}<{\centering} | p{0.85cm}<{\centering} | p{0.85cm}<{\centering}|}
		\hline
		\multicolumn{1}{|c|}{\multirow{2}{*}{\textbf{Backbones}}} &
		\multicolumn{2}{c|}{\textbf{DUTS}} &
		\multicolumn{2}{c|}{\textbf{ECSSD}} &
		\multicolumn{2}{c|}{\textbf{SOD}}  &
		\multicolumn{2}{c|}{\textbf{HKU-IS}} &
		\multicolumn{2}{c|}{\textbf{DUT-OMRON}} &
		\multicolumn{2}{c|}{\textbf{THUR 15K}}\\
		\cline{2-13}
		& $\mathrm{F}_\beta \uparrow$ & $\mathrm{MAE} \downarrow$ & $\mathrm{F}_\beta \uparrow$ & $\mathrm{MAE} \downarrow$ & $\mathrm{F}_\beta \uparrow$ & $\mathrm{MAE} \downarrow$ 
		& $\mathrm{F}_\beta \uparrow$ & $\mathrm{MAE} \downarrow$ & $\mathrm{F}_\beta \uparrow$ & $\mathrm{MAE} \downarrow$ & $\mathrm{F}_\beta \uparrow$ & $\mathrm{MAE} \downarrow$\\
		\hline
		\multicolumn{1}{|c|}{\multirow{2}{*}{VGG16}} & 0.823 & 0.059 & 0.897 & 0.095 & 0.795 & 0.165 & 0.896 & 0.082 & 0.743 & 0.098 & 0.748 & 0.110\\
		& 0.871  & 0.039 & 0.945 & 0.037 & 0.879 & 0.081 & 0.942 & 0.026 & 0.813 & 0.046 & 0.813 & 0.059\\
		\hline
		\multicolumn{1}{|c|}{\multirow{2}{*}{ResNet50}} & 0.835 & 0.056 & 0.914 & 0.087 & 0.824 & 0.135 & 0.919 & 0.075 & 0.763 & 0.085 & 0.769 & 0.094\\
		& 0.887  & 0.032 & 0.958 & 0.032 & 0.891 & 0.063 & 0.956 & 0.022 & 0.832 & 0.039 & 0.839 & 0.048\\
		\hline
		\multicolumn{1}{|c|}{\multirow{2}{*}{ResNeXt101}} & \textcolor{blue}{0.848} & \textcolor{blue}{0.053} & \textcolor{blue}{0.921} & \textcolor{blue}{0.079} & \textcolor{blue}{0.832} & \textcolor{blue}{0.128} & \textcolor{blue}{0.924} & \textcolor{blue}{0.073} & \textcolor{blue}{0.771} & \textcolor{blue}{0.080} & \textcolor{blue}{0.773} & \textcolor{blue}{0.089}\\
		& \textcolor{red}{0.891}  & \textcolor{red}{0.030} & \textcolor{red}{0.961} & \textcolor{red}{0.031} & \textcolor{red}{0.896} & \textcolor{red}{0.060} & \textcolor{red}{0.958} & \textcolor{red}{0.021} & \textcolor{red}{0.837} & \textcolor{red}{0.037} & \textcolor{red}{0.845} & \textcolor{red}{0.045}\\
		\hline
	\end{tabular}
	\caption{Precision of edges and regions with three different backbones, {\em i.e.}, VGG16, ResNet50 and ResNeXt101, in which the first and second row show the results of edges and regions, respectively. In particular, the best results of edges are denoted in blue and the best results of regions are indicated by red.}
	\label{tab_2}
	\vspace{-0.2cm}
\end{table*}

\textbf{Implementation.} We used Pytorch to implement our algorithm. The hardware environment is a PC with Intel Core CPUs 3.4 GHz, 32 GB memory and NVIDIA GTX 1080Ti GPU. The batch size is set to be $10$, the learning rate is initialized to be $0.01$ and decreased by a factor of $0.1$ at every two epochs. In the training process, we first randomly crop $300\times300$ from input images, then follow a random horizontal flipping for data augmentation. There are two hyper parameters in our weighted $\mathrm{L}2$ loss function, and we set $\rho = 3$ and $\sigma = 0.01$ in all the experiments.

\subsection{Ablation Study}
\textbf{Performances with different backbones.} Firstly, we want to evaluate the precision of edges and regions with different backbones, as shown in Table~\ref{tab_2}, in which we report the results of edges and regions in the first and second rows, respectively. In practice, the representation capability of ResNeXt101~\cite{Xie_Girshick_Piotr:2017} is stronger than that of ResNet50~\cite{He_Zhang_Ren:2016}, and the representation capability of ResNet50 is stronger than that of VGG16~\cite{Simonyan_Zisserman:2014}. As a result, our method achieves its best performance when the ResNeXt101 is chosen as backbone, which is about $2.0\%$ and $0.4\%$ higher in $\mathrm{F}_\beta$ and $0.9\%$, $0.2\%$ higher in $\mathrm{MAE}$ than that of VGG16 and ResNet50, respectively. Besides, we can see that the final performances achieved by the ResNet50 and ResNeXt101 are very close to each other, which indicates that our method is robust to the choice of backbone network.

\begin{figure}[t]
	\centering
	\begin{tabular}{c}
		\hspace{-0.2cm}
		\vspace{-0.1cm}
		\includegraphics[height = 2.5cm, width = 8.0cm]{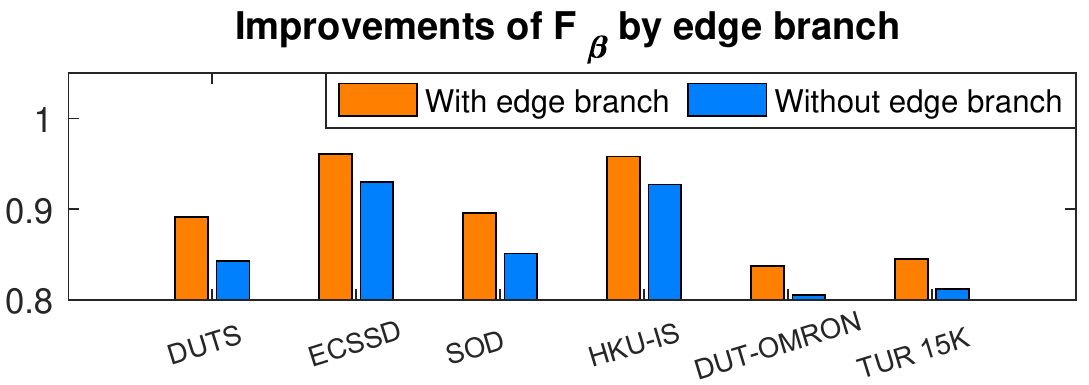} \\
		\hspace{-0.2cm}
		\includegraphics[height = 2.5cm, width = 8.0cm]{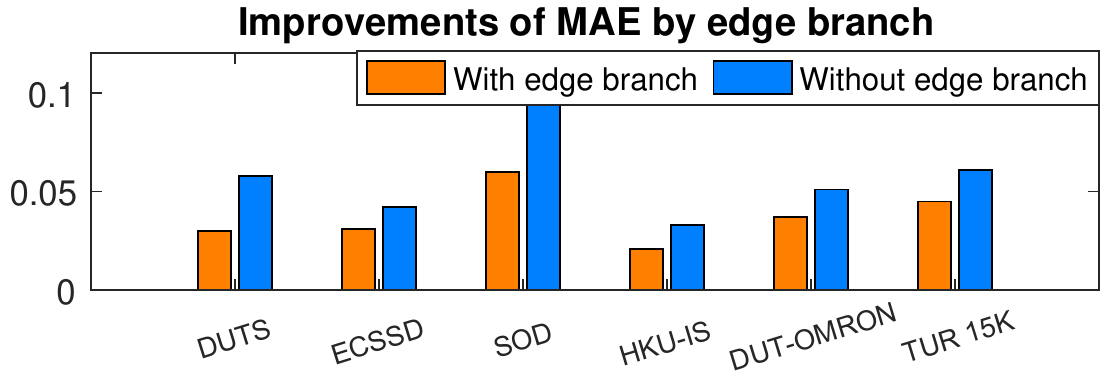} 
	\end{tabular}
	\caption{Improvements by the edge branch network on the six benchmark datasets, respectively.}
	\label{fig_5}
	\vspace{-0.2cm}
\end{figure}

\textbf{Performances at different stages.} Secondly, we want to evaluate the precision of edges and regions at different stages, as shown in Table~\ref{tab_3}, in which we choose the ResNeXt101 as backbone for simplicity. The first and second rows of Table~\ref{tab_3} show the results of edges and regions, respectively. Because our SE\textsuperscript{2}Net is a multi-stage network to refine the previous predictions, the quality of edges and regions consistently improved as the stage increases. For example, the results at stage 3 are about $2.0\%$ and $4.0\%$ higher in $\mathrm{F}_\beta$, and $1.0\%$ and $1.8\%$ higher in $\mathrm{MAE}$ than that at stage 2 and stage 1, respectively. The best performance of our method was reached at stage 3, in which the edge and region information are only fused twice in the training process. This leads to two conclusions: (1) The multi-stage network structure can consistently refine the salient maps of edges and regions; (2) The inference cost of our network is very small, because it only needs three stages to achieve its best performance.

\begin{figure}[t]
	\centering
	\begin{tabular}{c}
		\hspace{-0.2cm}
		\vspace{-0.1cm}
		\includegraphics[height = 2.5cm, width = 8.0cm]{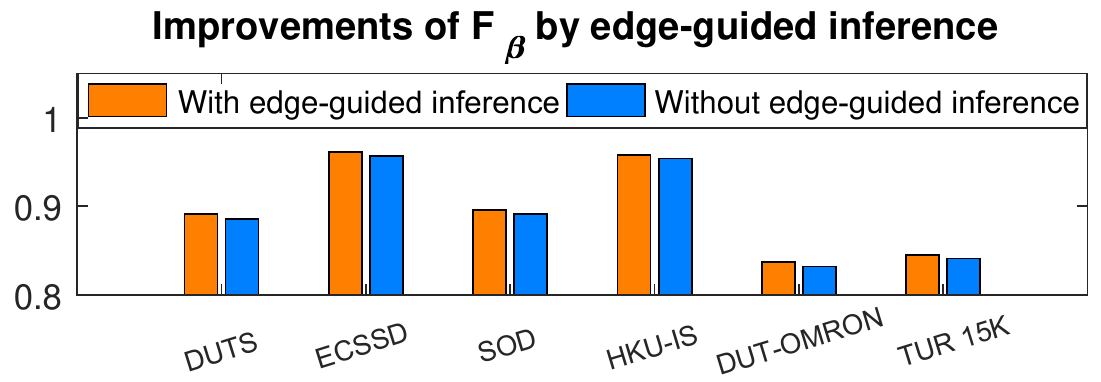} \\
		\hspace{-0.2cm}
		\includegraphics[height = 2.5cm, width = 8.0cm]{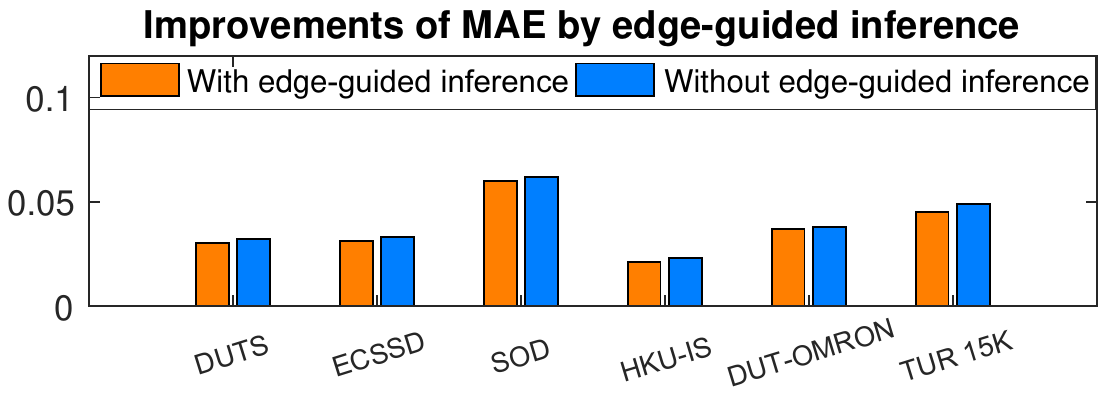} 
	\end{tabular}
	\caption{Improvements by the edge-guided inference algorithm on the six benchmark datasets, respectively}
	\label{fig_6}
	\vspace{-0.2cm}
\end{figure}

\begin{table*}[t]
	\centering
	\small
	\begin{tabular}{| p{1.5cm}<{\centering} | p{0.85cm}<{\centering} | p{0.85cm}<{\centering} | p{0.85cm}<{\centering} | p{0.85cm}<{\centering} | p{0.85cm}<{\centering} | p{0.85cm}<{\centering} | p{0.85cm}<{\centering} | p{0.85cm}<{\centering} | p{0.85cm}<{\centering} | p{0.85cm}<{\centering} | p{0.85cm}<{\centering} | p{0.85cm}<{\centering}|}
		\hline
		\multicolumn{1}{|c|}{\multirow{2}{*}{\textbf{Stages}}} &
		\multicolumn{2}{c|}{\textbf{DUTS}} &
		\multicolumn{2}{c|}{\textbf{ECSSD}} &
		\multicolumn{2}{c|}{\textbf{SOD}}  &
		\multicolumn{2}{c|}{\textbf{HKU-IS}} &
		\multicolumn{2}{c|}{\textbf{DUT-OMRON}} &
		\multicolumn{2}{c|}{\textbf{THUR 15K}}\\
		\cline{2-13}
		& $\mathrm{F}_\beta \uparrow$ & $\mathrm{MAE} \downarrow$ & $\mathrm{F}_\beta \uparrow$ & $\mathrm{MAE} \downarrow$ & $\mathrm{F}_\beta \uparrow$ & $\mathrm{MAE} \downarrow$ 
		& $\mathrm{F}_\beta \uparrow$ & $\mathrm{MAE} \downarrow$ & $\mathrm{F}_\beta \uparrow$ & $\mathrm{MAE} \downarrow$ & $\mathrm{F}_\beta \uparrow$ & $\mathrm{MAE} \downarrow$\\
		\hline
		\multicolumn{1}{|c|}{\multirow{2}{*}{$t=1$}} & 0.809 & 0.072 & 0.872 & 0.109 & 0.768 & 0.193 & 0.871 & 0.094 & 0.732 & 0.105 & 0.725 & 0.123\\
		& 0.851  & 0.048 & 0.931 & 0.043 & 0.852 & 0.111 & 0.920 & 0.035 & 0.787 & 0.062 & 0.786 & 0.074\\
		\hline
		\multicolumn{1}{|c|}{\multirow{2}{*}{$t=2$}} & 0.829 & 0.057 & 0.908 & 0.089 & 0.821 & 0.136 & 0.910 & 0.082 & 0.755 & 0.090 & 0.761 & 0.099\\
		& 0.872  & 0.040 & 0.947 & 0.036 & 0.884 & 0.078 & 0.947 & 0.024 & 0.817 & 0.045 & 0.821 & 0.055\\
		\hline
		\multicolumn{1}{|c|}{\multirow{2}{*}{$t=3$}} & 0.848 & \textcolor{blue}{0.053} & \textcolor{blue}{0.921} & \textcolor{blue}{0.079} & 0.832 & \textcolor{blue}{0.128} & \textcolor{blue}{0.924} & 0.073 & 0.771 & 0.080 & \textcolor{blue}{0.773} & \textcolor{blue}{0.089}\\
		& \textcolor{red}{0.891}  & \textcolor{red}{0.030} & 0.961 & \textcolor{red}{0.031} & \textcolor{red}{0.896} & \textcolor{red}{0.060} & \textcolor{red}{0.958} & \textcolor{red}{0.021} & 0.837 & \textcolor{red}{0.037} & \textcolor{red}{0.845} & \textcolor{red}{0.045}\\
		\hline
		\multicolumn{1}{|c|}{\multirow{2}{*}{$t=4$}} & \textcolor{blue}{0.849} & 0.054 & 0.918 & 0.081 & \textcolor{blue}{0.834} & 0.129 & 0.922 & \textcolor{blue}{0.071} & \textcolor{blue}{0.773} & \textcolor{blue}{0.078} & 0.770 & 0.090\\
		& 0.886 & 0.031 & \textcolor{red}{0.963} & 0.032 & 0.891 & 0.064 & 0.954 & 0.022 & \textcolor{red}{0.839} & \textcolor{red}{0.035} & 0.843 & 0.047 \\
		\hline
		\multicolumn{1}{|c|}{\multirow{2}{*}{$t=5$}} & 0.841 & 0.059 & 0.909 & 0.087 & 0.829 & 0.132 & 0.917 & 0.079 & 0.767 & 0.802 & 0.767 & 0.092\\
		& 0.881 & 0.035 & 0.957 & 0.038 & 0.886 & 0.077 & 0.949 & 0.023 & 0.831 & 0.041 & 0.838 & 0.049\\
		\hline
	\end{tabular}
	\caption{Results of our method with at different stages, in which we choose the ResNeXt101 as backbone. In particular, the first and second row show the results of edges and regions, respectively. Besides, the best results of edges are denoted in blue and the best results of regions are indicated by red.}
	\label{tab_3}
	\vspace{-0.2cm}
\end{table*}

\begin{figure*}[t]
	\centering
	\begin{tabular}{c}
		\hspace{-0.3cm}
		\includegraphics[height = 8.0cm, width = 17.5cm]{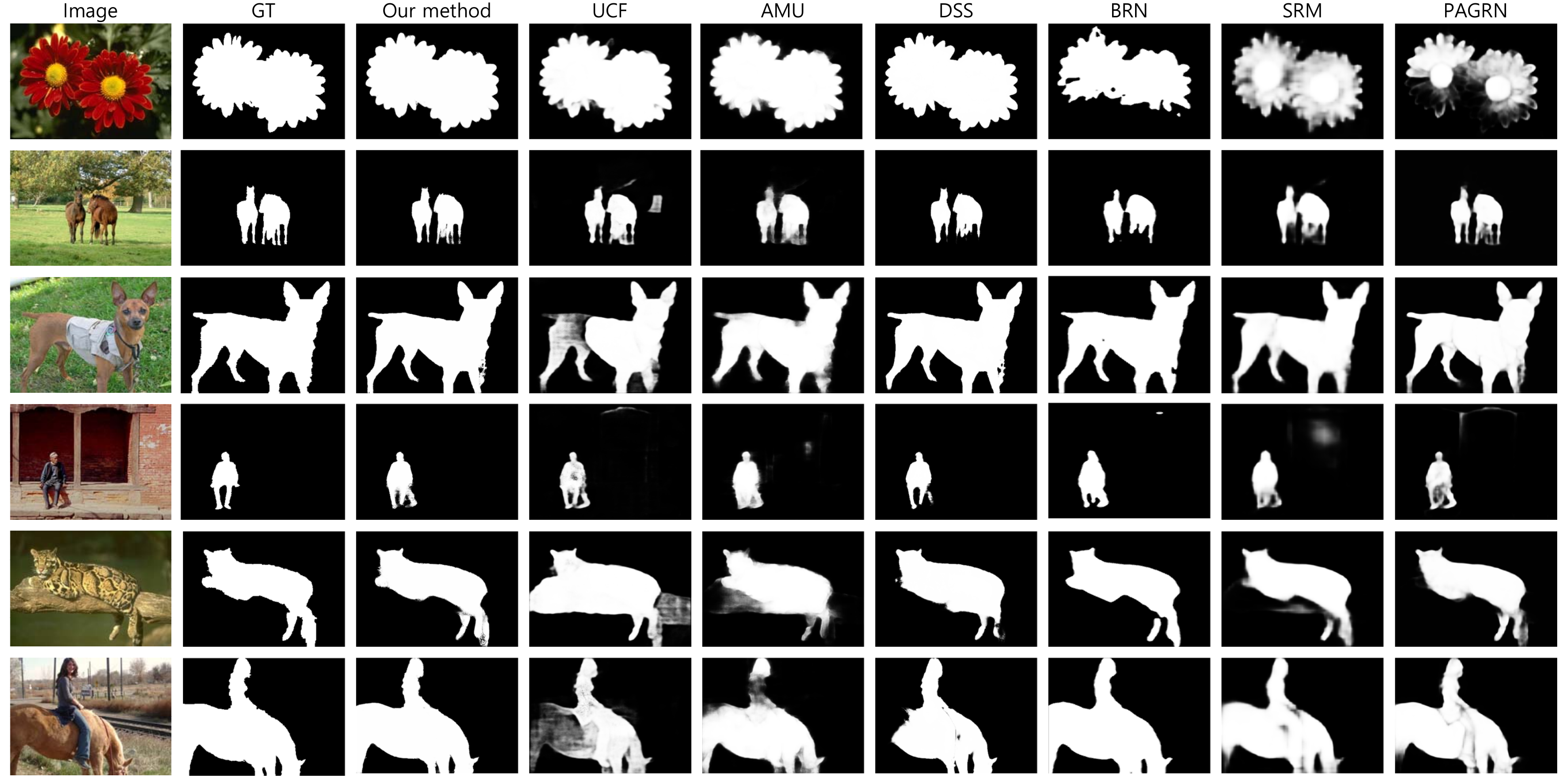}
	\end{tabular}
	\caption{Visualization of the resulting salient maps of regions by our method, UCF~\cite{Zhang_Wang_Lu:2017}, AMU~\cite{Zhang_Wang_Lu:2017}, DSS~\cite{Hou_Cheng_Hu:2017}, BRN~\cite{Wang_Zhang_Wang:2018}, SRM~\cite{Wang_Borji_Zhang:2017} and PAGRN~\cite{Zhang_Wang_Qi:2018}, respectively.}
	\label{fig_7}
	\vspace{-0.2cm}
\end{figure*}

\begin{figure}[t]
	\centering
	\begin{tabular}{c}
		\hspace{-0.2cm}
		\includegraphics[height = 4.0cm, width = 8.0cm]{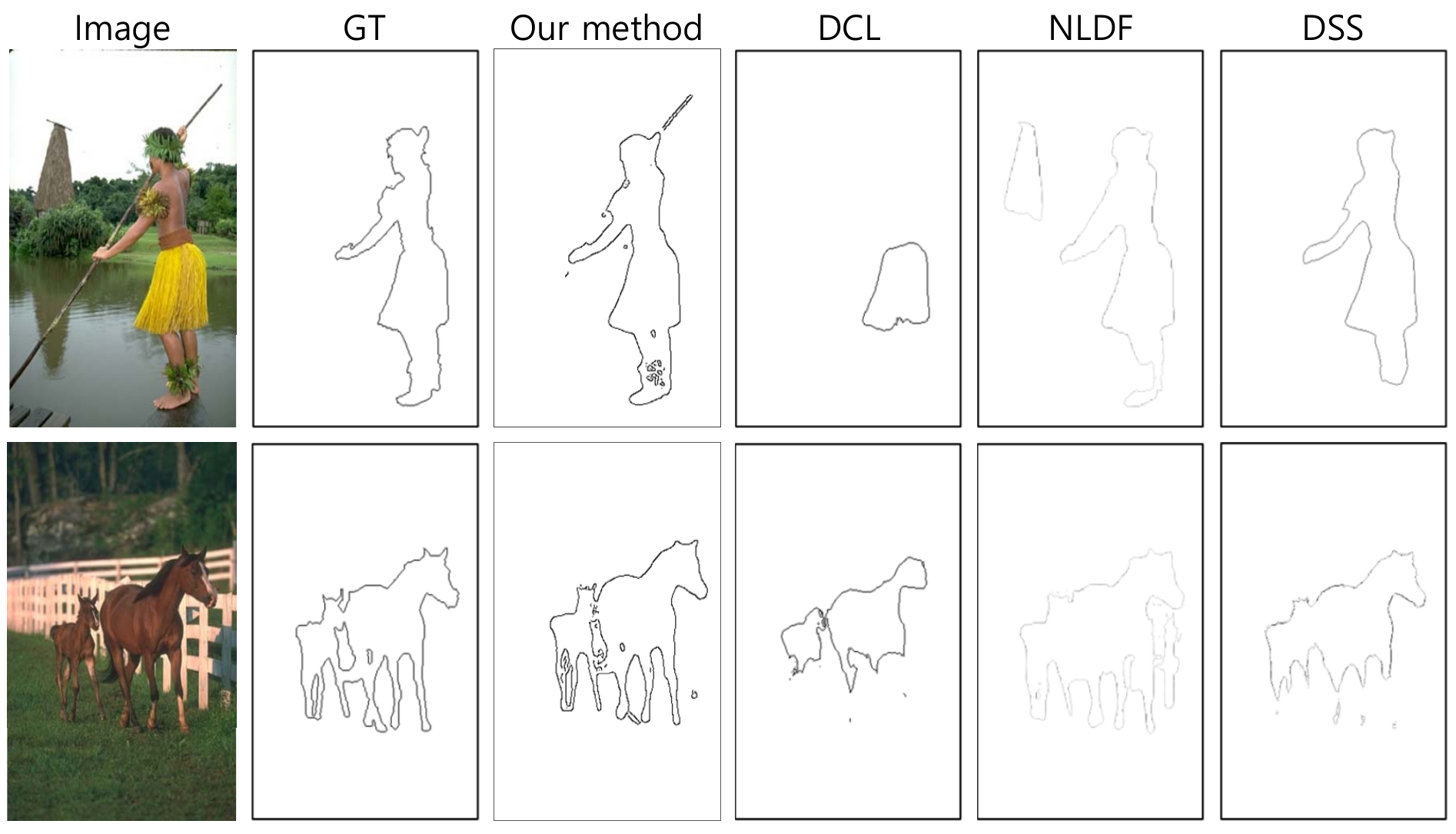}
	\end{tabular}
	\caption{Visualization of the resulting salient maps of edges by our method, DCL~\cite{Li_Yu:2016}, NLDF~\cite{Luo_Mishra_Achkar:2017} and DSS~\cite{Hou_Cheng_Hu:2017}, respectively.}
	\label{fig_8}
	\vspace{-0.2cm}
\end{figure}

\textbf{Improvements by edge branch network.} Thirdly, we want to evaluate how much the edge branch network contributes to the final results. For consistency, we still use the ResNeXt101 as backbone and take three stages to conduct two sets of experiments: the first experiment uses both the edge branch network and the region branch network, while the second experiment only uses the region branch network. The results are shown in Figure~\ref{fig_5}, from which we can see that both the $\mathrm{F}_\beta$ and $\mathrm{MAE}$ were significantly improved by introducing the edge branch network. These results have verified that it is very important to apply the edge branch network to help the region branch network in dealing with the blurred edges.

\textbf{Improvements by edge-guided inference algorithm.} Fourthly, we want to evaluate how much the edge-guided inference algorithm can improve the final results. Because our method can jointly estimate the salient maps of edges and regions, it is very convenient to use our edge-guided inference algorithm to refine the salient maps of regions. The results are shown in Figure~\ref{fig_6}, from which we can see that the edge-guided inference algorithm consistently improved the final $\mathrm{F}_\beta$ and $\mathrm{MAE}$ on each dataset. Besides, we also notice that the improvements are relatively small as compared with the ones achieved by our edge branch network. Because the edge branch network has already handled some weak edges in the training process, it will be very hard to significantly improve the resulting results.

\begin{table*}[t]
	\centering
	\small
	\begin{tabular}{| p{1.5cm} | p{0.85cm}<{\centering} | p{0.85cm}<{\centering} | p{0.85cm}<{\centering} | p{0.85cm}<{\centering} | p{0.85cm}<{\centering} | p{0.85cm}<{\centering} | p{0.85cm}<{\centering} | p{0.85cm}<{\centering} | p{0.85cm}<{\centering} | p{0.85cm}<{\centering} | p{0.85cm}<{\centering} | p{0.85cm}<{\centering}|}
		\hline
		\multicolumn{1}{|c|}{\multirow{2}{*}{\textbf{Methods}}} &
		\multicolumn{2}{c|}{\textbf{DUTS}} &
		\multicolumn{2}{c|}{\textbf{ECSSD}} &
		\multicolumn{2}{c|}{\textbf{SOD}}  &
		\multicolumn{2}{c|}{\textbf{HKU-IS}} &
		\multicolumn{2}{c|}{\textbf{DUT-OMRON}} &
		\multicolumn{2}{c|}{\textbf{THUR 15K}}\\
		\cline{2-13}
		& $\mathrm{F}_\beta \uparrow$ & $\mathrm{MAE} \downarrow$ & $\mathrm{F}_\beta \uparrow$ & $\mathrm{MAE} \downarrow$ & $\mathrm{F}_\beta \uparrow$ & $\mathrm{MAE} \downarrow$ 
		& $\mathrm{F}_\beta \uparrow$ & $\mathrm{MAE} \downarrow$ & $\mathrm{F}_\beta \uparrow$ & $\mathrm{MAE} \downarrow$ & $\mathrm{F}_\beta \uparrow$ & $\mathrm{MAE} \downarrow$\\
		\hline
		\multicolumn{13}{|c|}{\textbf{VGG16}}\\
		\hline
		RFCN~\cite{Wang_Wang_Lu:2016} & 0.782 & 0.089 & 0.896 & 0.097 & 0.802 & 0.161 & 0.892 & 0.080 & 0.738 & 0.095 & 0.754 & 0.100\\
		NLDF~\cite{Luo_Mishra_Achkar:2017} & 0.806 & 0.065 & 0.902 & 0.066 & 0.837 & 0.123 & 0.902 & 0.048 & 0.753 & 0.080 & 0.762 & 0.080\\
		PiCA~\cite{Liu_Han_Yang:2018} & 0.837 & 0.054 & 0.923 & 0.049 & 0.836 & 0.102 & 0.916 & 0.042 & 0.766 & 0.068 & 0.783 & 0.083\\
		C2S~\cite{Li_Yang_Cheng:2018}  & 0.811 & 0.062 & 0.907 & 0.057 & 0.819 & 0.122 & 0.898 & 0.046 & 0.759 & 0.072 & 0.775 & 0.083\\
		RAS~\cite{Chen_Tan_Wang:2018}  & 0.831 & 0.059 & 0.916 & 0.058 & 0.847 & 0.123 & 0.913 & 0.045 & 0.785 & 0.063 & 0.772 & 0.075\\
		HCA~\cite{Liu_Qiu_Zhang:2018}  & 0.858 & 0.044 & 0.933 & 0.042 & 0.856 & 0.108 & 0.927 & 0.031 & 0.791 & 0.057 & 0.788 & 0.071\\
		\hline
		SE\textsuperscript{2}Net & \textbf{0.871}  & \textbf{0.039} & \textbf{0.945} & \textbf{0.037} & \textbf{0.879} & \textbf{0.081} & \textbf{0.942} & \textbf{0.026} & \textbf{0.813} & \textbf{0.046} & \textbf{0.813} & \textbf{0.059}\\
		\hline
		\multicolumn{13}{|c|}{\textbf{ReSNet50}}\\
		\hline
		SRM~\cite{Wang_Borji_Zhang:2017}  & 0.826 & 0.059 & 0.914 & 0.056 & 0.840 & 0.126 & 0.906 & 0.046 & 0.769 & 0.069 & 0.778 & 0.077\\
		BRN~\cite{Wang_Zhang_Wang:2018}  & 0.827 & 0.050 & 0.919 & 0.043 & 0.843 & 0.103 & 0.910 & 0.036 & 0.774 & 0.062 & 0.769 & 0.076\\
		RAS~\cite{Chen_Tan_Wang:2018}  & 0.857  & 0.052 & 0.921 & 0.045 & 0.847 & 0.101 & 0.912 & 0.039 & 0.781 & 0.069 & 0.779 & 0.078\\
		PiCA~\cite{Liu_Han_Yang:2018} & 0.853 & 0.050 & 0.929 & 0.049 & 0.852 & 0.103 & 0.917 & 0.043 & 0.789 & 0.065 & 0.788 & 0.081\\
		R\textsuperscript{3}Net~\cite{Deng_Hu_Zhu:2018}  & 0.861 & 0.048 & 0.932 & 0.050 & 0.860 & 0.102 & 0.923 & 0.036 & 0.795 & 0.063 & 0.793  & 0.063\\
		HCA~\cite{Liu_Qiu_Zhang:2018}  & 0.875 & 0.040 & 0.942 & 0.036 & 0.865 & 0.099 & 0.934 & 0.029 & 0.819 & 0.054 & 0.796 & 0.069\\
		\hline
		SE\textsuperscript{2}Net  & \textbf{0.887}  & \textbf{0.032} & \textbf{0.958} & \textbf{0.032} & \textbf{0.891} & \textbf{0.063} & \textbf{0.956} & \textbf{0.022} & \textbf{0.832} & \textbf{0.039} & \textbf{0.839} & \textbf{0.048}\\
		\hline
	\end{tabular}
	\caption{Comparison of SE\textsuperscript{2}Net with the state-of-the-art approaches on the six datasets, respectively. In particular, we compare with the competitors on multiple backbone networks. The best results are denoted in bold black.}
	\label{tab_4}
	\vspace{-0.2cm}
\end{table*}

\subsection{Comparisons}
\textbf{Comparison with state-of-the-arts.} Firstly, we want to compare our method with the state-of-the-art methods in salient object detection. The results are shown in Table~\ref{tab_4}, in which we compared our  SE\textsuperscript{2}Net with the state-of-the-art approaches with different backbone networks. The previous best results were achieved by HCA~\cite{Liu_Qiu_Zhang:2018}, and our method outperformed it about $1.2\%$ in $\mathrm{F}_\beta$ and $1.5\%$ in $\mathrm{MAE}$, and $1.2\%$ in $\mathrm{F}_\beta$ and $1.8\%$ in $\mathrm{MAE}$ when the VGG16 and ResNet50 are chosen as backbone, respectively. Besides, we reproduced the results of R\textsuperscript{3}Net~\cite{Deng_Hu_Zhu:2018} with their source codes, because they also estimate the salient maps in a multi-stage way. It can be seen that our method outperforms them for about $2.6\%$ in $\mathrm{F}_\beta$ and $1.6\%$ in $\mathrm{MAE}$, with the help of edge branch network.

\begin{figure}[t]
	\centering
	\begin{tabular}{c}
		\hspace{-0.2cm}
		\includegraphics[height = 2.0cm, width = 7.8cm]{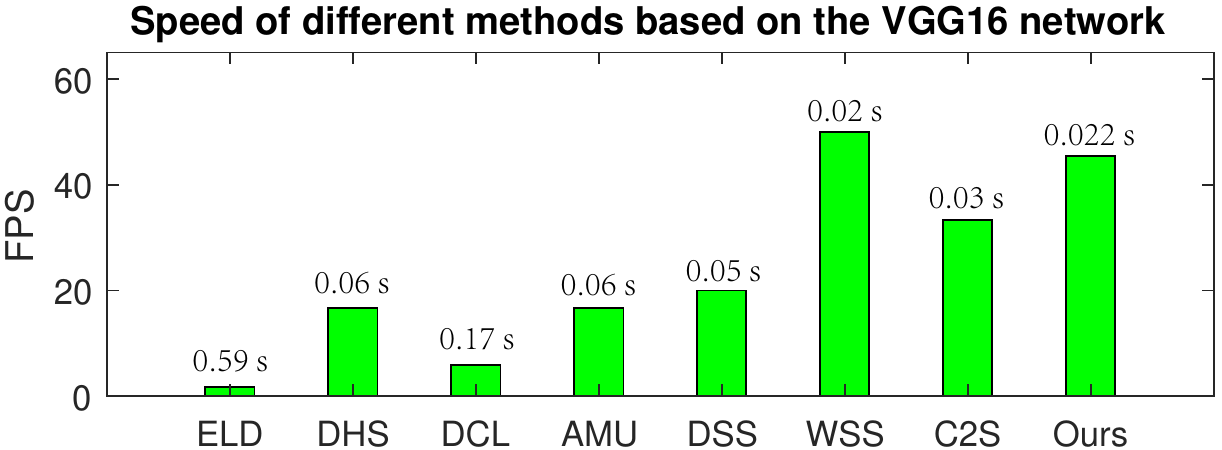}
	\end{tabular}
	\caption{Speed of different methods, including the DSS~\cite{Hou_Cheng_Hu:2017}, ELD~\cite{Lee_Tai_Kim:2016}, DHS~\cite{Liu_Han:2016}, DCL~\cite{Li_Yu:2016}, AMU~\cite{Zhang_Wang_Lu:2017}, WSS~\cite{Wang_Lu_Wang:2017}, C2S~\cite{Li_Yang_Cheng:2018} and our method.}
	\label{fig_9}
	\vspace{-0.2cm}
\end{figure}

\textbf{Visualization of salient maps.} Secondly, we want to compare the resulting salient maps of edges and regions of our method with the state-of-the-art approaches. The results are shown in Figure~\ref{fig_7} and Figure~\ref{fig_8}, in which we can see that: (1) The region maps generated by our method are more effective to preserve the edges of object, and (2) The edge maps generated by our method are more semantic in suppressing the false positives in background. In fact, this is the key reason why we try to learn the salient maps of edges and regions in a unified network.

\textbf{Comparison with edge-based methods.} Thirdly, we want to compare our method with the edge-based approaches. For simplicity, we evaluate all these methods on the ESSCD dataset, and the results are shown in Table~\ref{tab_5}. From the results, we can see that our method has achieved the best results. The reason comes in two aspects: (1) Our method can learn and fuse the edge and region information in each stage; (2) Our method can learn the high-quality edge maps in a supervised manner. 

\textbf{Speed of training and testing.} Fourthly, we want to compare the speed of our method with the state-of-the-art approaches. The results are shown in Figure~\ref{fig_9}, in which all the methods used VGG16 as backbone for fair comparison. In particular, our method takes about 2 hours to complete the training process in 6000 iterations. Besides, it only takes 0.022 second to produce a saliency map for a $300 \times 300$ input image, which is very competitive as compared with the other methods.

\begin{table}[t]
	\centering
	\small
	\begin{tabular}{|l| p{0.9cm}<{\centering} | p{0.9cm}<{\centering} | p{0.9cm}<{\centering} | p{0.9cm}<{\centering} | p{0.9cm}<{\centering} |}
		\hline
		Methods & MSR & AMU & DSS & RFCN &  Ours\\
		\hline
		$\mathrm{F}_\beta \uparrow$   & 0.913  & 0.864 & 0.915 & 0.834& \textbf{0.958}\\
		\hline
		$\mathrm{MAE} \downarrow$ & 0.054 & 0.059 & 0.052 & - & \textbf{0.032}\\
		\hline
	\end{tabular}
	\caption{Comparison with these edge-based approaches on the ESSCD dataset, {\em i.e.}, the MSR~\cite{Li_Xie_Lin:2017}, AMU~\cite{Zhang_Wang_Lu:2017}, DSS~\cite{Hou_Cheng_Hu:2017}, RFCN~\cite{Wang_Wang_Lu:2016} and our method, in which `-' means they don't report the result.}
	\label{tab_5}
	\vspace{-0.2cm}
\end{table}

\vspace{0.25cm}
\section{Conclusion}
In this paper, we proposed a simple yet effective Siamese Edge-Enhancement Network to preserve the edge structure for salient object detection. Firstly, a novel multi-stage siamese network is designed to parallelly estimate the salient maps of edges and regions from the low-level and high-level features. As a result, the predicted regions become more accurate by enhancing the responses at edges, and the predicted edges become more semantic by suppressing the false positives in background. Secondly, an edge-guided inference algorithm is designed to further improve the resulting masks along the predicted edges. Extensive experimental results on several benchmark datasets have shown that our method is superior than most of the state-of-the-art approaches.

\clearpage
{\small
\bibliographystyle{ieee}
\bibliography{SE2Net}
}

\end{document}